\documentclass[runningheads]{llncs}
\usepackage{amsmath}
\usepackage{cite} 
\usepackage[T1]{fontenc}
\usepackage{cite}
\usepackage{multirow}

\usepackage{textcomp}
\usepackage{xcolor}
\usepackage{float}
\usepackage{comment}
\usepackage{subcaption}
\usepackage{hyperref}
\usepackage{url}
\usepackage{multirow}
\usepackage{booktabs}
\usepackage{array} 

\usepackage{graphicx}
\usepackage[linesnumbered]{algorithm2e}
\RestyleAlgo{ruled}
\SetKwComment{Comment}{// }{}
\SetKwInput{KwInput}{Input}
\SetKwInput{KwOutput}{Output}
\SetKwInput{KwInitialization}{Initialization}
\SetKw{Continue}{continue}

\begin{document}
\title{Energy-Efficient Path Planning with Multi-Location Object Pickup for Mobile Robots on Uneven Terrain}

\author{Faiza Babakano  \and
Ahmed Fahmin \and
Bojie Shen  \and \\
Muhammad Aamir Cheema \and 
Isma Farah Siddiqui  
}
\authorrunning{F. Babakano et al.}
\titlerunning{Time-Efficient Path Planning Algorithm for Mobile Robots}
\institute{
Faculty of Information Technology, Monash University, Melbourne, Australia
\\
\email{\{faiza.babakano,ahmed.fahmin,bojie1.shen,aamir.cheema,\\ismahfarah.siddiqui\}@monash.edu}
}
\maketitle
\begin{abstract}
 Autonomous Mobile Robots (AMRs) operate on battery power, making energy efficiency a critical consideration particularly in outdoor environments where terrain variations affect energy consumption. While prior research has primarily focused on computing energy-efficient paths from a source to a destination, these approaches often overlook practical scenarios where a robot needs to pick up an object en route—an action that can significantly impact energy consumption due to changes in payload. This paper introduces the Object-Pickup Minimum Energy Path Problem (OMEPP), which addresses energy-efficient route planning for Autonomous Mobile Robots (AMRs) required to pick up an object from one of the many possible locations and take it to a destination. 
To address the OMEPP problem, we first introduce a baseline algorithm that employs the Z* algorithm, a variant of A* tailored for energy-efficient routing, to iteratively visit each pickup point. While this approach guarantees optimality, it suffers from high computational cost due to repeated search efforts at each pickup location. To mitigate this inefficiency, we propose a concurrent PCPD search that manages multiple Z* searches simultaneously across all pickup points. Central to our solution is the Payload-Constrained Path Database (PCPD), an extension of the Compressed Path Database (CPD), a state-of-the-art technique for fast shortest path computation, that incorporates payload constraints. 
We further demonstrate that PCPD significantly reduces branching factors during search, leading to improved overall performance.
Although the concurrent PCPD search may produce slightly suboptimal solutions, extensive experiments on real-world datasets demonstrate that it achieves near-optimal performance while being one to two orders of magnitude faster than the baseline algorithm derived from existing methods.

\end{abstract}


\section{Introduction}\label{sec:intro}

Autonomous Mobile Robots (AMRs) are advanced systems capable of navigating and performing tasks independently through control system and real-time decision-making. 
Although AMRs are predominantly utilized in structured indoor environments, there is a growing imperative to extend their deployment to unstructured and dynamic outdoor settings, such as forests, agricultural fields, and disaster zones, which present significant challenges, including uneven terrain that necessitate advanced and robust path planning techniques.
Outdoor deployment of AMRs enables transformative applications such as environmental monitoring~\cite{nguyen2021mobile}, autonomous delivery~\cite{alverhed2024autonomous}, and search-and-rescue missions~\cite{vacariu2004multiagent}, 
underscoring the need for robust, adaptable 
robotic systems capable of operating safely and efficiently in outdoor conditions.

 \begin{figure*}[t]
 \centering
 \includegraphics[width= 0.8\columnwidth]{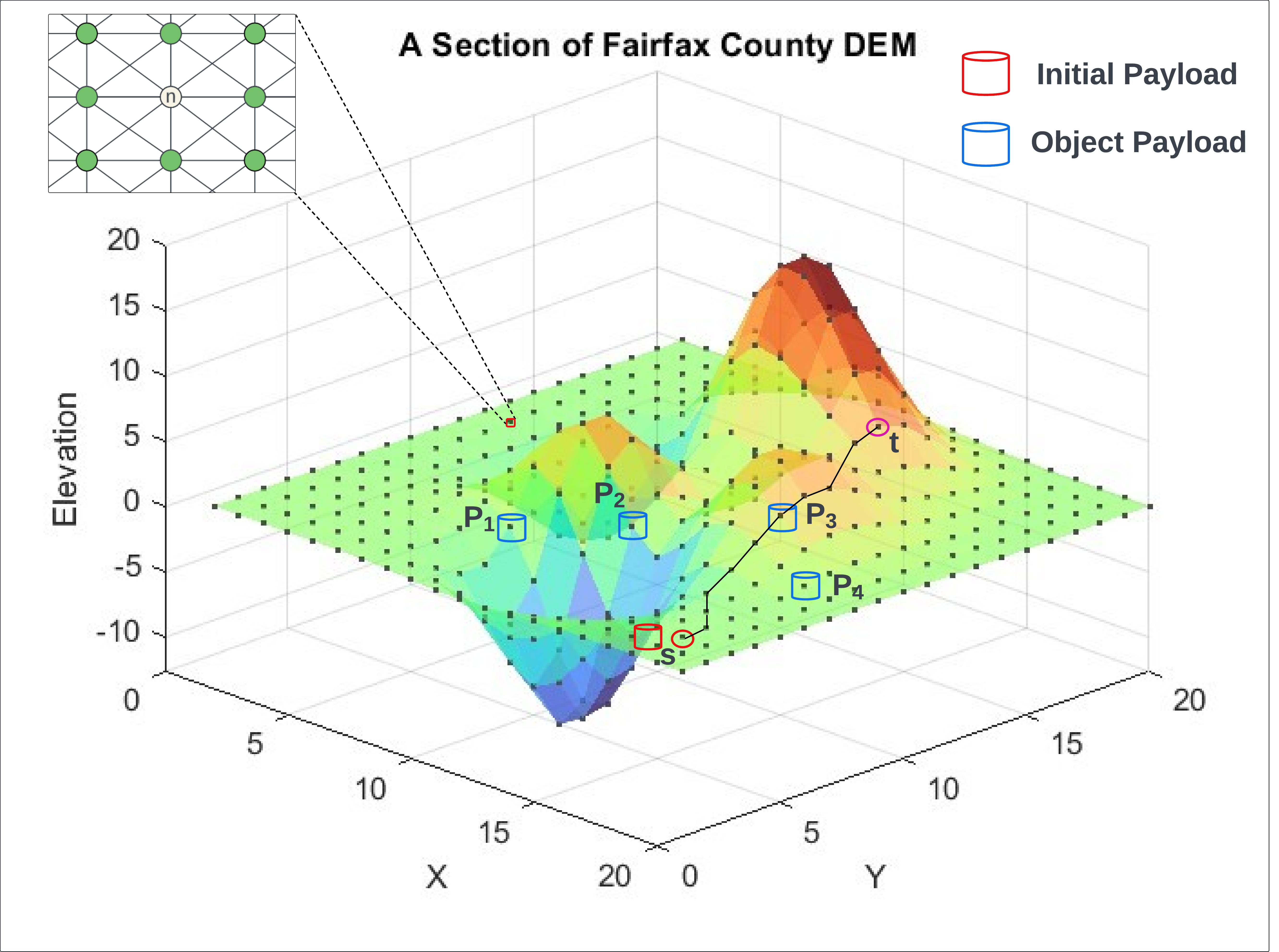}
 \caption{
 A toy example of the Object-Pickup Minimum Energy Path Problem (OMEPP). Pickup locations $\{p_1, p_2, p_3, p_4\}$ are shown in blue. The optimal path, which minimizes energy consumption, is illustrated in the figure, where the robot starts at $s$, picks up an object from the pickup location $p_3$, and proceeds to the target $t$. 
 }
 \label{fig:mash_graph}
 \end{figure*}

Path planning is a fundamental aspect of AMR navigation and has attracted substantial attention in recent years. Traditionally, much of the literature has focused on computing paths over uneven terrain between a start and target location, optimizing for the shortest geodesic distance~\cite{wei2024efficient} or adhering to kinodynamic constraints~\cite{BabakanoFSC24}. However, there is a growing interest in energy-efficient path planning, particularly in outdoor environments where terrain features such as hills and ridges can significantly increase energy consumption~\cite{wei2020energy}. This shift is further driven by the rising cost of energy, which has intensified efforts toward designing energy-aware navigation systems~\cite{liu2013minimizing}. Despite these advancements, many existing approaches still assume that AMRs need only travel between a start and goal location, without considering the execution of intermediate tasks.
In practice, however, AMRs are often required to interact with the environment by picking up or delivering objects along their paths. For example, as demonstrated in~\cite{jeong2024dynamic}, delivery robots may collect packages from designated pickup points and transport them to customers. Similarly, in domains such as search and rescue, agriculture, or manufacturing, AMRs may be tasked with transporting supplies, gathering soil samples, or moving workpieces, respectively. These tasks typically involve objects that are distributed across the environment, adding additional complexity to the path planning problem.
Consider the example of a large construction site with uneven terrain where an autonomous robot is required to deliver a heavy tool to a workstation. The same tool is available at several storage spots across the site. The robot needs to plan a route that includes picking up the tool from one location and delivering it to the destination in the most energy-efficient way possible.

To the best of our knowledge, we are the first to study the \underline{O}bject-Pickup \underline{M}inimum \underline{E}nergy \underline{P}ath \underline{P}roblem (OMEPP) on outdoor terrain. In this problem, a mobile robot must travel from a given source to a destination while picking up an object from one of several possible pickup locations along the way.
 OMEPP aims to determine both the optimal location from where to pick up the object and the corresponding energy-efficient path, such that the total energy consumption of the journey is minimized. 
Fig.~\ref{fig:mash_graph} illustrates a simplified instance of the OMEPP problem, which contains multiple pickup points $\{p_1, p_2, p_3, p_4\}$ distributed across uneven terrain.
To keep the problem general, we assume the robot must carry essential materials for the mission. It starts at location 
$s$ with an initial payload $\rho_{\text{init}}$, collects an object from one of the designated pickup points, and then ascends the mountain to reach the target location $t$.
Planning a path in this context is non-trivial, as the robot must account for the initial payload $\rho_{\text{init}}$ as well as the additional payload $\rho_{\text{obj}}$ after retrieving an object from a selected pickup point $p_i$. The payload of a robot can influence both the maximum slope the robot can ascend and its overall energy consumption. Therefore, selecting the appropriate pickup point $p_i$ and determining an energy-efficient path from $s$ to $p_i$, and subsequently from $p_i$ to $t$, are of critical importance. In this example, the optimal solution involves selecting pickup point $p_3$ for delivery to the target $t$, as shown in the figure.

Despite the practical importance of object pickup, recent studies on energy-efficient path planning for mobile robots~\cite{ganganath2015constraint, saad2019energy, choi2012global, wei2019air} typically assume fixed start and goal locations, overlooking intermediate tasks such as object pickup. 

To efficiently solve the OMEPP problem, we begin by designing a baseline algorithm that leverages the existing Z* algorithm, a variant of A* that employs a novel admissible heuristic to estimate energy costs for computing energy-efficient paths. The baseline algorithm iteratively considers each pickup point $p_i$ and runs two independent instances of Z* to compute the optimal energy-efficient paths from $s$ to $p_i$, and from $p_i$ to $t$. While this approach guarantees optimality, it suffers from inefficiency due to redundant search efforts at each pickup point.
To address this limitation, we propose a concurrent PCPD search that manages multiple Z* searches across different pickup points simultaneously. However, the concurrent search may generate up to 
$m \times n$ successors,  where $m$ and $n$ denote the number of neighbors considered in the searches from $s$ to $p_i$ and from 
$p_i$ to $t$, respectively. To mitigate this combinatorial expansion, we draw inspiration from the Compressed Path Database (CPD) approach \cite{fast_cpd, botea11}. A CPD is a technique that avoids traditional state-space search by relying on precomputed data to retrieve shortest paths efficiently. Instead of computing paths at runtime, a CPD functions as an oracle: given a start state 
$s$ and target $t$, the query CPD$[s,t]$ immediately returns the first edge on the optimal path from $s$ to $t$. By recursively following this first-move guidance, the full shortest path can be easily reconstructed~\cite{s-gppc-14}.
Accordingly, we adapt the CPD framework to reduce the number of successor nodes generated during concurrent search. In general, our approach employs the following strategies:

\begin{itemize}
\item We propose a concurrent search framework that employs a two-level best-first strategy to manage multiple $Z^*$ searches across different pickup points simultaneously. The high-level search maintains a global priority queue to select the most promising pickup point, while each low-level search uses a dedicated local queue to resume a concurrent $Z^*$ search. These low-level searches simultaneously expand paths from the source $s$ to a pickup point $p_i$, and from $p_i$ to the target $t$.

\item We further extend the Compressed Path Database (CPD) by introducing the Payload-Constrained Path Database (PCPD), which precomputes and stores the first moves along minimum-energy paths for varying payload levels. The payload range is partitioned into $n$ equal-sized buckets $\{\rho_1, \dots, \rho_n\}$, and for each bucket $\rho_i$, we construct a separate $\text{CPD}(\rho_i)$ that stores the first move on the minimum-energy path under the corresponding payload constraint.

\item We propose a PCPD-based successor generation method. Given the total payload $\rho$ on the $s$ (resp. $p_i$) side of the search, we select two CPDs: $\text{CPD}(\rho_{\text{lower}})$ and $\text{CPD}(\rho_{\text{upper}})$, where $\rho_{\text{lower}} \leq \rho \leq \rho_{\text{upper}}$. Each CPD suggests a first move toward $p_i$ (resp. $t$); if both moves are identical, only one successor is generated. Otherwise, we take the union of valid moves, resulting in at most two successors per side. Thus, each low-level search generates up to four successors in total.

\end{itemize}

We describe the concurrent PCPD search algorithm and the construction process for Payload-Constrained Path Databases (PCPDs).
To evaluate our approach, we conduct a series of experiments on real-world uneven terrain. Compared to the baseline algorithm, our method achieves significantly faster performance, with one to two orders of magnitude, while maintaining near-optimal path quality.

\section{Related Work}

 Energy-efficient path planning is crucial for mobile robots given that they are often battery-powered. Although batteries can be recharged, minimizing energy consumption reduces downtime and increases the overall efficiency of robotic operations. Shortest-path algorithms based on geodesic distance have been extensively studied, however, they do not necessarily produce the most energy-efficient routes \cite{wei2020energy}. As a result, a growing body of research focuses on developing methods that prioritize energy optimization over purely geodesic shortest paths.

One of the earliest efforts in this direction was by Rowe and Ross \cite{rowe1990optimal}, who introduced an energy cost function based on gravitational and frictional forces acting on the robot to plan energy-efficient routes in 3D space. Subsequent studies, including \cite{1391019}, \cite{choi2012global}, \cite{saad2019energy}, \cite{ganganath2014finding}, and \cite{ganganath2015constraint} incorporated this energy cost function with classical graph-based shortest path algorithms like A* and Dijsktra to find optimal energy paths. Choi \textit{et. al.}\cite{choi2012global} employed A* on a grid-based graph generated from simulated terrain to plan energy-efficient paths. However, the heuristic functions developed in the study were inadmissible and failed to generate alternate paths for very steep slopes. To address these limitations, \cite{ganganath2015constraint} developed a heuristic energy-cost function that is both admissible and consistent. This enabled the generation of zigzag-like paths that avoid cells with slopes exceeding the robot's climbing capability. The proposed Z* algorithm operates on the scale of hundreds of seconds which is slow for real-life applications. 
Building from\cite{ganganath2015constraint}, a modified A* was used in \cite{ganganath2016multiobjective} to generate a set of non-dominated routes between two locations in uneven terrain. The multiobjective search algorithm was developed to avoid the trade-off between the shortest and energy-optimal routes This allows planners to select a path that balances length and energy efficiency. 

Zakharov \textit{et.al.}\cite{zakharov2020energy} derived a mathematical formula that uses slope and height differences between neighboring cells of a grid to identify non-traversable areas. The non-traversable cells are marked as obstacle and then classical A* was used to find the shortest path assuming all energy consuming grids are marked as obstacle. However, the actual energy savings from this algorithm cannot be precisely measured, which makes it difficult to evaluate its accuracy.

On the contrary, numerous studies used other approaches in finding energy efficient path.  Aksamentov \textit{et.al.} \cite{aksamentov2020approach} processed aerial images  using an R-CNN neural network to detect static obstacles. A 3D model of the environment is then constructed and used by a their algorithm to plan the robot's route. Kyau \textit{et.al.} \cite{kyaw2022energy} worked specifically on reconfigurable robots. The energy efficient path was generated  using a sampling based method called BRT* while  \cite{saad2019energy}, introduced a new composite routing metric that combines energy consumption and travel distance, leading to a modified version of Dijkstra's algorithm for path planning.
Sun and Reif \cite{1391019} used triangular meshes extracted from the Digital Elevation Model (DEM) and converted them into a weighted graph. By combining terrain discretization with the BUSHWHACK algorithm and an efficient approximation method, the authors were able to compute energy-minimizing paths. However, the approximation of the terrain into flat surfaces degrades the accuracy of the generated paths.

Different studies have attempted to solve the energy-efficient path planning problem as discussed above; however, to the best of our knowledge, these studies assume that the initial payload remains constant from the start node $s$ to the target node $t$, which is not the case in most real-world applications. In addition, studies using graph-based approaches are relatively slow for practical use.

\section{Preliminaries}
\label{sec:Preliminaries}
\subsection{Problem Formulation}

Advances in geographic information systems have enabled the widespread availability of high-resolution Digital Elevation Models (DEMs), offering precise representations of terrain surface elevation across diverse regions. The data is organized as a set of nodes $N$, where each node $n \in N$ has three-dimensional coordinates $(n.x, n.y, n.z)$. Nodes are linked to their eight neighboring nodes, with a minimum spacing of 1 meter between them.
Fig~\ref{fig:mash_graph} shows an example of DEM extracted from Fairfax County.

We model the uneven terrain as a directed graph $G = (V, E, D, \Theta)$, consistent with the structure of the original data. The graph comprises a set of vertices $V$, where each vertex corresponds to a node $n \in N$. Each edge $e_{v_i v_j} \in E$ represents a directed connection from vertex $v_i$ to vertex $v_j$. The functions $d \in D$ and $\phi \in \Theta$ map each edge $e_{v_i v_j}$ to a non-negative weight $d(v_i, v_j)$ and a slope value $\phi(v_i, v_j)$, respectively. Here, $d(v_i, v_j)$ represents the distance required to traverse from $v_i$ to $v_j$, while $\phi(v_i, v_j)$ represents the slope between the two vertices.

Following prior work, we utilize the energy consumption model proposed in \cite{ganganath2015constraint} to accurately estimate the energy consumption of a mobile robot during path traversal. This model accounts for the two primary resistive forces acting on the robot: gravitational and frictional forces.
Assuming the robot has mass $m$, moves at a constant velocity 
$cv$, and has a maximum available power $P_{\max}$, its energy consumption is significantly influenced by the total payload 
$\rho$, which consists of the initial payload
$\rho_{\text{init}}$ and any additional load $\rho_{\text{obj}}$ picked up during traversal.
Given the total payload $\rho$, the energy consumption
for moving between two nodes $v_i$ and $v_j$ on an inclined surface is defined as:
\begin{equation}
    k(\rho, v_i, v_j) = (\rho+m) g s(v_i, v_j)\left[\mu \cos \theta(v_i, v_j) + \sin \theta(v_i, v_j)\right]
\end{equation}
Here, the variable $g$ is the gravitational acceleration, $ s(v_i, v_j)$ is the Euclidean distance between the two nodes, $\mu$ is the coefficient of friction, and $\theta(v_i, v_j)$ represents the slope angle between the nodes.
This formulation accounts for both frictional resistance and the gravitational effort required to ascend or descend slopes. However, 
An increased payload can affect the robot's motion capabilities and impose constraints on the slopes it can feasibly traverse. The slope constraints are defined as follows:

\begin{itemize}

    \item The maximum uphill slope angle, denoted as $\phi_\gamma$, is constrained by the available traction force. The maximum traction force, $F_{\text{max}}$, can be approximated as $P_{\text{max}} / (cv)$. The slope limit due to traction is given by:

    \begin{equation}
    \phi_f = \sin^{-1}\left(\frac{F_{\text{max}}}{(\rho + m) g \sqrt{\mu^2 + 1}}\right) - \tan^{-1}(\mu)
    \label{eq:maxAngle}
    \end{equation}
    
    In addition, traction is limited by the static friction coefficient $\mu_s$ between the vehicle and the surface. Anisotropic traction loss occurs when the slope angle exceeds a critical value $\phi_s$, defined as:

    \begin{equation}
    \phi_s = \tan^{-1}(\mu_s - \mu)
    \end{equation}

Taking both constraints into account, the maximum permissible uphill slope angle is defined as:  $\phi_\gamma = min(\phi_f, \phi_s)$.

    \item The critical downhill slope angle $\phi_c$ is the steepest slope the robot can descend without requiring active braking (i.e., the gravitational component does not exceed resistive friction). It is given by:
    \begin{equation}
        \phi_c = \tan^{-1}(\mu)
    \end{equation}
\end{itemize}

To incorporate these constraints, the energy cost for a robot traveling from $v_i$ to $v_j$ can be summarized as:
  
\begin{equation}
     k(\rho,v_i, v_j) = 
    \begin{cases}
        \infty, & \text{if } \theta(v_i, v_j) > \phi_\gamma \\
        0, & \text{if } \theta(v_i, v_j) < \phi_c \\
        (\rho + m) g s(v_i, v_j)[\mu \cos \theta(v_i, v_j) \\ \hspace{2.1cm} + \sin \theta(v_i, v_j)], & \text{otherwise}
    \end{cases}
    \label{eq:enery_consumption}
\end{equation}

A \emph{path} $P$ from a start node $s \in V$ to a target node $t \in V$ is defined as a sequence of vertices $\langle v_0, v_1, v_2, \ldots, v_{q-1}, v_q \rangle$, where $v_0 = s$, $v_q = t$, and each edge $e_{v_i v_{i+1}} \in E$ for all $0 \leq i < q$.  
The \emph{energy cost} of the path is given by
$|P|_E = \sum_{i=0}^{q-1} k( \rho, v_i, v_{i+1})$
where $k(\rho, v_i, v_{i+1})$ denotes the energy required for a robot with total payload $\rho$ to traverse from $v_i$ to $v_{i+1}$.
We denote the \emph{minimum-energy path} from $s$ to $t$ by $ep(\rho, s, t)$, and the corresponding \emph{minimum energy distance} by $ed(\rho, s, t)$.
We use the notation $\oplus$ to join paths together, assuming the two endpoints are identical. That is, given paths $\langle v_1, \ldots, v_n \rangle$ and $\langle v_1', v_2', \ldots, v_m' \rangle$, the expression $\langle v_1, \ldots, v_n \rangle \oplus \langle v_1', v_2', \ldots, v_m' \rangle$ generates the path $\langle v_1, \ldots, v_n, v_2', \ldots, v_m' \rangle$, assuming $v_n = v_1'$.

The robot is tasked with picking up objects of the same type, each with an identical payload denoted by $\rho_{\text{obj}}$. For example, these could be sample containers or equipment packages in an outdoor exploration or environmental monitoring scenario. The uneven terrain contains multiple such objects, located at a set of pick-up points $\{p_1, p_2, \dotsc, p_n\} \subseteq V$, where each $p_i$ corresponds to a vertex in the graph $V$. Next, we formally define the Object-Pickup Minimum Energy Path Problem (OMEPP):

\textbf{Object-Pickup Minimum Energy Path Problem (OMEPP):}  
Given a source node $s \in V$, a target node $t \in V$, and a set of pick-up points $\{p_1, p_2, \dotsc, p_n\} \subset V$, where each pick-up point contains an object with payload $\rho_{\text{obj}}$. OMEPP seeks to determine the optimal path that minimizes the total energy cost for picking up an object from start and reaching the target. Formally, it finds the entire path that minimizes:
\[
P = \arg\min_{\{p_1, p_2, \dotsc, p_n\}} \left( ep(\rho_{\text{init}}, s, p_i) + ep(\rho_{\text{init}} + \rho_{\text{obj}}, p_i, t) \right),
\]

\subsection{Compressed Path Database(CPDs)}
The Compressed Path Database (CPD) is a state-of-the-art algorithm for global pathfinding in grid maps \cite{s-gppc-14}. The key advantage of CPD over traditional shortest path algorithms is its ability to precompute the shortest path information, which allows for efficient path retrieval during query processing. CPD has a broad range of applications across various domains, including Euclidean space \cite{EPS}, road networks \cite{CHCPD}, and uneven terrain \cite{babakano2025time}. In board stroke, the CPD operates in two distinct stages:

\begin{itemize}
    \item \textbf{Offline Preprocessing Phase:}
    In this stage, CPD precomputes the first move on the shortest path from each source node $s$ to every destination node $t$ using a modified Dijkstra algorithm. The result is a first-move table, denoted $T$, which typically requires $O(n^2)$ space, where $n$ is the number of nodes in the graph. To reduce space complexity, Run-Length Encoding (RLE) is employed to efficiently compress sequences of repeated values, with each row $T(s)$ sorted using a Depth-First Search (DFS) traversal as recommended in \cite{fast_cpd}. Although the offline precomputation phase is time-intensive, each Dijkstra search is independent and can be executed in parallel, achieving a speedup proportional to the number of available processors.
    \item \textbf{Online Stage:} 
    During the online stage, the precomputed data structure cpd is used to efficiently reconstruct the optimal path from a source node $s$ to a target node $t$. The function $cpd[s,t]$, which returns the first move from $s$ toward $t$, can be efficiently implemented via binary search on the compressed first-move row T(s). Given this, the complete path is obtained recursively: starting from \(s\), we define the next node as \(v_1 = \texttt{cpd}[s, t]\), then \(v_2 = \texttt{cpd}[v_1, t]\), and so on, until \(v_k = t\). This sequence of nodes \(\langle s = v_0, v_1, v_2, \ldots, v_k = t \rangle\) forms the reconstructed path.


\end{itemize}

 \begin{figure}[t]
 \centering
 \includegraphics[width=0.7\columnwidth]{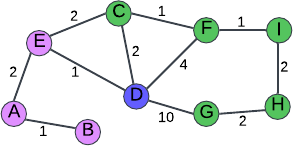}
 \caption{\textbf{From the source D, the optimal first move to any node colored purple (resp.green) is E (resp.C).}}
   \captionsetup{belowskip=0pt} 
 \label{fig:grapnh}
 \end{figure}
 
\begin{example}
\textit{Consider the graph shown in Fig.~\ref{fig:grapnh}. The first moves table for nodes D, A, and I is shown in Table 1. Note that each CPD[s,s] is assigned a wildcard symbol "*", which can be combined with any run, as we never need to look up a first move from s to itself. The first moves table is ordered by running a DFS from vertex D. CPD compresses the first moves table via RLE. For example, the row D: *, E, E, E, C, C, C, C, C can be compressed into two runs [1:E] and [5:C]
]. The compressed RLE string for each row is given as ([1:E],[5:C]), ([1:E],[4:B],[5:E]), and ([1:F],[8:H]).}
\end{example} 

\setlength{\tabcolsep}{8pt} 

\captionsetup[table]{skip=12pt} 

\renewcommand{\arraystretch}{1.2} 

\begin{table}[t]
\centering
\caption{First Moves for D, A and I for example of Fig.~\ref{fig:grapnh} }
\bfseries 
\begin{tabular}{c|c|c|c|c|c|c|c|c|c}
\hline
\noalign{\hrule height 0.5 pt} 
Ordering & D & E & A & B & C & F & I & H & G \\
\hline
\hline
D & * & E & E & E & C & C & C & C & C \\
\hline
A & E & E & * & B & E & E & E & E & E \\
\hline
I & F & F & F & F & F & F & * & H & H \\
\hline
\noalign{\hrule height 0.5 pt} 
\end{tabular}
\label{tab:example}
\end{table}

\section{Methodology}
\label{sec:analysis}
In this section, we propose efficient algorithms to solve the Object-Pickup Minimum Energy Path Problem (OMEPP). As no existing work directly addresses this problem, we first introduce a baseline algorithm using the Z* algorithm~\cite{ganganath2015constraint}, an A* variant with a modified admissible and consistent heuristic. This approach visits each pickup point iteratively to compute the optimal path. While Z* improves search efficiency, it remains slow due to repeated searches at each pickup.
To overcome this, we propose the concurrent PCPD search, which explores multiple pickup paths simultaneously. We also introduce the Payload-constrained Compressed Path Database (PCPD), an extension of CPD that incorporates varying payload information into the first-move table, reducing branching factors in order to efficiently solve OMEPP.

\begin{algorithm}[t]
\SetAlgoNoEnd
\caption{Baseline Algorithm}
\label{alg:Z*-algorithm}
\KwIn{Source node $s \in V$, Target node $t \in V$, Pickup points $\{p_1, p_2, \dotsc, p_n\} \subset V$, Initial payload $\rho_{\text{init}}$, Object payload $\rho_{\text{obj}}$}
\KwOut{Path with total minimum energy cost}
\KwInitialization{$E_{\text{min}} = \infty$; $P_{\text{best}} = \text{null}$;}
\For{each $p_i$ in $\{p_1, p_2, \dotsc, p_n\}$ \label{algo1:line:for_each}}{
    $ep(\rho_{\text{init}}, s, p_i) = \text{Z}^* Search(\rho_{\text{init}}, s , p_i)$\;\label{algo1:line:first_path}
    $ep(\rho_{\text{init}} + \rho_{\text{obj}}, p_i, t) = \text{Z}^* Search(\rho_{\text{init}} + \rho_{\text{obj}},p_i , t)$\; \label{algo1:line:second_path}
    $E_{\text{total}} = |ep(\rho_{\text{init}},s,p_i)|_E + |ep(\rho_{\text{init}} + \rho_{\text{obj}},p_i,t)|_E$\; \label{algo1:line:energy_sum}
    \If{$E_{\text{total}} < E_{\text{min}}$ \label{algo1:line:energy_min1}}{
        $E_{\text{min}} = E_{\text{total}}$\;
        \label{algo1:line:energy_min2}
        $P_{\text{best}}= ep(\rho_{\text{init}}, s,p_i) \oplus ep(\rho_{\text{init}} + \rho_{\text{obj}}, p_i,t);$ 
        \label{algo1:line:path_update}
        
    }
}
\textbf{Return} $P_{\text{best}}$ with total energy cost $E_{\text{min}}$; \label{algo1:line:return}
\end{algorithm}

\subsection{Baseline Algorithm}
To solve the OMEPP, we first design a baseline algorithm that utilizes the Z* algorithm to compute energy-efficient paths. The core idea of the baseline algorithm is to iteratively compute the minimum-energy path from a source node 
$s$ to a target node $t$, while passing through each pickup point in the candidate set $\{p_1, p_2, \dotsc, p_n\}$. The pseudocode for the baseline algorithm is provided in Algorithm~\ref{alg:Z*-algorithm}.
Next, we begin by presenting the main idea of the algorithm, and a detailed explanation of the Z* algorithm will follow.

The algorithm initializes the minimum energy cost $ E_{\text{min}}$ to infinity and the best path $P_{\text{best}}$ to null. It then iterates over each candidate pickup point  $p_i \in \{p_1, p_2, \dotsc, p_n\}$ (line~\ref{algo1:line:for_each}). For each candidate, two point-to-point energy-optimal paths are computed using the $ \text{Z}^*$ search. The first path connects $s$ to $p_i$
under the initial payload $\rho_{\text{init}}$ (line~\ref{algo1:line:first_path}), while the second connects $p_i$ to $t$ under the increased payload $\rho_{\text{init}} + \rho_{\text{obj}}$ (line~\ref{algo1:line:second_path}). The total energy cost is calculated as the sum of the energy costs of these two segments (line~\ref{algo1:line:energy_sum}).
If the resulting cost is less than the current minimum $ E_{\text{min}}$, the algorithm updates both the minimum cost and the corresponding path $P$, which is formed by concatenating the two segments (lines~\ref{algo1:line:energy_min1} -~\ref{algo1:line:path_update}). After all pickup points have been evaluated, the algorithm returns the path $P_{\text{best}}$ with the lowest total energy cost (line~\ref{algo1:line:return}).

\subsubsection{Z* algorithm~\cite{ganganath2015constraint}}
 is a variant of the well-known A* search algorithm, designed to improve performance in energy efficient pathfinding scenarios where traditional heuristics such as euclidean heuristic may fall short. 
Similar to A*, which finds optimal paths by combining the cost to reach a node (i.e., the g-value) with an admissible heuristic estimate to the goal (i.e., the h-value), Z* builds upon this foundation by introducing a novel energy-cost heuristic function. This function takes into account the varying slope values a robot can traverse relative to its total payload $\rho$, as described in Equation~(\ref{eq:enery_consumption}). Specifically, the energy heuristic is given by:

\begin{equation}
\label{eq:heurCost}
           \textit{h}(\rho,v_c, t) =
\begin{cases} 
    \frac{(\rho + m) g\Delta(v_c,t)}{sin\theta_\gamma}[\mu cos\theta_\gamma \\ \hspace{2cm}+ sin\theta_\gamma], & \text{if } \theta(v_c, t) > \phi_\gamma  \\
    0,  & 
  \text{if } \theta(v_c, t) < \phi_c  \\
    (\rho + m) g s(v_c, t)[\mu cos\theta(v_c, t) \\ 
   \hspace{2cm} + sin\theta(v_c, t)] & otherwise
\end{cases}
\end{equation} 

where $\Delta(v_c,t)$ denote the elevation difference between the current node $v_c$ and the goal node $t$, defined as $\Delta(v_c,t) = v_c.z - t.z$. The slope $\theta(v_c,t)$ is computed by drawing a virtual straight line between $v_c$ and $t$. The heuristic energy-cost function $\textit{h}(\rho, v_c, t)$ estimates the energy required for a robot to travel from $v_c$ to $t$, taking into account the following three scenarios:
(i) When the slope $\theta(v_c, t)$ exceeds the robot’s maximum allowable angle $\theta_\gamma$ (i.e., $\theta(v_c, t) > \theta_\gamma$), the robot must follow a zigzag path, and the energy cost is approximated using the detoured trajectory.
(ii) When the slope is below $\theta_c$ (i.e., $\theta(v_c, t) < \theta_c$), the robot falls in the braking range, and the energy cost is negligible or zero.
(iii) When the slope is between $\phi_c$ and $ \phi_\gamma$ (i.e., $\theta_c < \theta(v_c, t) > \theta_\gamma$), the robot is able to traverse the angles less than the critical impermissible angle, and the energy cost is computed as $h(\rho, v_c, t) = k(\rho, v_c, t)$. For further details, refer to the original Z* algorithm paper~\cite{ganganath2015constraint}.

\subsection{Concurrent PCPD Search}
The baseline algorithm requires running two independent Z* searches for each pickup point $p_i$: one from the source $s$ to $p_i$, and another from $p_i$ to the destination $t$. This approach can be inefficient due to redundant search efforts around $s$ and 
$t$, especially when the number of pickup locations is large. To address this limitation, we propose the Concurrent PCPD Search, which maintains multiple Z* searches, one for each pickup point, allowing simultaneous exploration. Unlike the standard Z* search, we extend the concept of the Compressed Path Database (CPD) to introduce the Payload-Constrained Compressed Path Database (PCPD). PCPD effectively reduces the branching factor of standard Z* search by leveraging precomputed path data under payload constraints.

\begin{algorithm}[t]
\SetAlgoNoEnd
\caption{Concurrent PCPD Search}
\label{alg:concurrent_pcpd_search}
\KwIn{Source $s$, Target $t$, Pickup points $\{p_1, \dotsc, p_n\}$, Initial payload $\rho_{\text{init}}$, Object payload $\rho_{\text{obj}}$, PCPD module}
\KwOut{Optimal path from $s$ to $t$ via one pickup point that minimizes total energy consumption}

\KwInitialization{$GQ \gets \emptyset$; initialize each $CQ[p_i \dots p_n] \gets \emptyset$}

\For{each pickup point $p_i$ in $\{p_1, \dotsc, p_n\}$}{
    Estimate cost $f = h(\rho_{\text{init}}, s, p_i) + h(\rho_{\text{init}}+ \rho_{\text{obj}}, p_i, t)$ \; \label{algo2:line:computeh}
    Insert node $\{s\rightarrow p_i, p_i\rightarrow t\}$ with cost $f$ into $CQ[p_i]$ \;
    \label{algo2:line:initCQ}
    Insert node $p_i$ with cost $f$ into $GQ$ \;
    \label{algo2:line:initGQ}
}

\While{$GQ$ is not empty}{
    SearchNode$(p_i, f) =$ $GQ$.pop() \;     \label{algo2:line:GQpop}
    \If{$CQ[i]$ is not empty}{
        SearchNode$(\{v_i \rightarrow p_i, v_j \rightarrow t\}, f) =$ $CQ[p_i]$.pop() \;  \label{algo2:line:CQpop}
    
        \If{$v_i = p_i$ and $v_j = t$}{
            \textbf{Return} optimal path with total cost $f$; 
            \label{algo2:line:returnPath}
        }
        $\mathcal{S} =$ generate successors of $\{v_i \rightarrow p_i, v_j \rightarrow t\}$ using payloads $\rho_{\text{init}}$ and $\rho_{\text{init}} + \rho_{\text{obj}}$ via PCPD \; \label{algo2:line:generateSuccessor}
        
        \For{each successors $\{v_i' \rightarrow p_i, v_j' \rightarrow t\} \in \mathcal{S}$}{
            Insert node $\{v_i' \rightarrow p_i, v_j' \rightarrow t\}$ with the updated cost $f'$ into $CQ[p_i]$\;
            \label{algo2:line:insertCQ}        }
        Insert node $p_i$ with cost $CQ[p_i].top()$ into $GQ$ \;
        \label{algo2:line:insertGQ}
    }
}
\textbf{Return} null;  \label{algo2:line:null}
\end{algorithm}

\subsubsection{Managing the Z* search Concurrently} 
We presents a \textit{two-level best-first search framework} for computing an energy-efficient path from a source node $s$ to a destination $t$, via a pickup point from a given set $\{p_1, \dotsc, p_n\}$. 
The high-level search maintains a global priority queue to select the most promising pickup point $p_i$, while each low-level search uses a dedicated local queue to continue a concurrent Z* search.
A key aspect of the approach is the use of a successor generation technique that utilized Payload-constrained Compressed Path Database (PCPD) to significantly reduce the branching factor (will be explained in a later section).
Algorithm~\ref{alg:concurrent_pcpd_search} shows the pseudocode of the proposed algorithm. Specifically, the algorithm relies on two types of priority queues to structure the search:
\begin{itemize}
    \item \textbf{Global Queue (GQ):} A priority queue that stores the current best estimated cost of reaching the goal via each pickup point. Each search node is a tuple $(p_i, f)$, where $p_i$ is a pickup point and $f$ is the cost associated with the top node in its corresponding child queue $CQ[p_i]$. This queue uses $f$ as key value to prioritize the pickup options and ensures that the most promising pickup path is always explored next.
    
    \item \textbf{Child Queues (CQ):} 
     A set of localized priority queues, $CQ[p_1], \dotsc, CQ[p_n]$, is maintained, with each queue corresponding to a specific pickup point $p_i$. Each queue performs a best-first search that simultaneously expands paths from the source $s$ to a pickup point $p_i$, and from $p_i$ to the target $t$. A search node is represented as a tuple $(\{v_i \rightarrow p_i, v_j \rightarrow t\}, f)$, where $v_i$ is a vertex along the partial path from $s$ to $p_i$, and $v_j$ is a vertex along the partial path from $p_i$ to $t$. The estimated cost of each component is given by $f_{v_i} = g(\rho_{\text{init}}, s, v_i) + h(\rho_{\text{init}},v_i, p_i)$ and $f_{v_j} = g(\rho_{\text{init}}+ \rho_{\text{obj}}, p_i, v_j) + h(\rho_{\text{init}}+ \rho_{\text{obj}},v_j, t)$, where $g$ denotes the actual cost and $h$ denote the heuristic estimate. The total cost is then $f = f_{v_i} + f_{v_j}$. Similar to standard A* search, each node maintains a pointer to its predecessor to support path reconstruction. All child queues operate independently, using $f$ as the key value to prioritize search nodes.

\end{itemize}

The search begins by initializing $GQ$ and all $CQ[p_i]$ as empty. For each pickup point $p_i$, an initial cost estimate is computed using the heuristic in Equation~\ref{eq:heurCost} (line~\ref{algo2:line:computeh}), and the node $({s \rightarrow p_i, p_i \rightarrow t}, f)$ is inserted into $CQ[p_i]$ and $(p_i, f)$ into $GQ$ (lines~\ref{algo2:line:initCQ}–\ref{algo2:line:initGQ}).

The main loop runs until $GQ$ is empty. In each iteration, the pickup point $(p_i, f)$ with the lowest cost is extracted (line~\ref{algo2:line:GQpop}), and if $CQ[p_i]$ is non-empty, the lowest-cost node $({v_i \rightarrow p_i, v_j \rightarrow t}, f)$ is expanded (line~\ref{algo2:line:CQpop}). If this node reaches the goal ${p_i, t}$, the path is reconstructed by backtracking (line~\ref{algo2:line:returnPath}); otherwise, successors are generated with payloads $\rho_{\text{init}}$ and $\rho_{\text{init}} + \rho_{\text{obj}}$ (line~\ref{algo2:line:generateSuccessor}). Since up to $m \times n$ successors may result, the Payload-Constrained Compressed Path Database (PCPD) is used to reduce branching.

For each successor ${v_i' \rightarrow p_i, v_j' \rightarrow t}$, the updated cost $f'$ is computed from $g(\cdot)$ and $h(\cdot)$ values, and the node is inserted into $CQ[p_i]$ (line~\ref{algo2:line:insertCQ}). After expansion, $GQ$ is updated with $(p_i, CQ[p_i].\text{top}())$ to reprioritize pickup points (line~\ref{algo2:line:insertGQ}). If $GQ$ is exhausted without reaching the goal, the algorithm terminates with failure (line~\ref{algo2:line:null}).

\subsubsection{Reducing Branching Factor} 
\label{subsec:pcpd}
As previously noted, each time Algorithm~\ref{alg:concurrent_pcpd_search} expands a search node $(\{v_i \rightarrow p_i, v_j \rightarrow t\}, f)$, it may generate successors up to $m \times n$ successors, potentially causing a combinatorial explosion that significantly degrades the performance of the proposed algorithm. To mitigate this issue, we aim to reduce the branching factor by selectively generating only high-quality successors. To this end, we extend the concept of the Compressed Path Database (CPD) precompute the first move along the optimal minimal-energy path under varying payload constraints. This results in the Payload-Constrained Compressed Path Database (PCPD). By leveraging the precomputed first-move data from the PCPD, we can efficiently reduce the number of successors generated to at most four.

\textbf{Payload-Constrained Compressed Path Database (PCPD).}
In the OMEPP problem, the total payload $\rho$ of a robot is a critical parameter that can substantially influence the planned path. Specifically, $\rho$ (i) determines the maximum slope the robot can traverse, and (ii) directly impacts energy consumption, as defined in Equation~(\ref{eq:enery_consumption}).
Motivated by this observation, we extend the concept of the Compressed Path Database (CPD) to introduce the Payload-Constrained Compressed Path Database (PCPD). Unlike the CPD, which stores the first move on the shortest distance path, PCPD takes a given payload $\rho'$ as input and stores the first move along the minimal-energy path that respects the associated constraints.
In the following, we describe the construction of PCPD in detail:

Given a payload constraint $\rho'$, we first determine the maximum slope value $\theta'$ that a robot can traverse while carrying the total payload $\rho'$ using Equation~(\ref{eq:maxAngle}). To construct a CPD that encodes the first move on the optimal (i.e., minimum energy) path under payload $\rho'$, we modify Dijkstra’s algorithm to minimize energy consumption and restrict traversal to edges with slopes no greater than $\theta'$. Specifically, for each outgoing edge $e_{v_i,v_j}$, the algorithm checks whether the slope $\theta(v_i, v_j)$ satisfies $\theta(v_i, v_j) \leq \theta'$. For those edges that meet this condition, it computes the energy consumption $k(\rho', v_i, v_j)$ using Equation~(\ref{eq:enery_consumption}) to update the tentative cost (i.e., $g$-value). The remainder of the procedure follows the standard Dijkstra framework.
After running this modified Dijkstra search from each source node $s \in V$ to compute the first-move table $T$, we apply RLE to compress $T$ and obtain the final CPD, denoted as CPD($\rho'$). This CPD can then be used to generate feasible (though potentially suboptimal) paths from any source $s$ to any target $t$ for a robot with any payload $\rho \leq \rho'$.

 Recall that when Algorithm~\ref{alg:concurrent_pcpd_search} generates successors for a search node $(\{v_i \rightarrow p_i, v_j \rightarrow t\}, f)$, it must account for the variable payloads $\rho_{\text{init}}$ and $\rho_{\text{init}} + \rho_{\text{obj}}$ at $v_i$ and $v_j$, respectively. Ideally, we would construct a CPD$(\rho)$ for every possible payload value $\rho$, based on standard robot configurations (e.g., $\rho_{\text{init}}$) and common object weights (e.g., $\rho_{\text{obj}}$), allowing us to always extract the truly optimal first move. However, generating such a large number of CPDs is both time-consuming and storage-intensive. To address this, we discretize the payload range into $n$ equal-sized buckets $\varrho = \{ \rho_1, \dots, \rho_n \}$. For each representative payload value $\rho_i$, we construct a corresponding CPD$(\rho_i)$ that stores first move on the minimum-energy path under payload $\rho_i$. 

\textbf{Successors Generation.} With the PCPD builds prepared, we now describe how to efficiently generate successors for a search node $(\{v_i \rightarrow p_i, v_j \rightarrow t\}, f)$ with respect to the pickup point $p_i$. Given the initial payload $\rho_{\text{init}}$ at location $v_i$, we first identify two CPDs: CPD($\rho_{lower}$) and CPD($\rho_{upper}$). Here, $\rho_{lower}$ is the largest payload $\rho_i \in \varrho$ such that $\rho_i \leq \rho_{\text{init}}$, and $\rho_{upper}$ is the smallest payload $\rho_i \in \varrho$ such that $\rho_i \geq \rho_{\text{init}}$.
For each of the selected CPDs:

\begin{itemize}
    \item CPD($\rho_{lower}$): 
    We extract the first move toward $p_i$ from CPD ($\rho_{lower}$)[$v_i, p_i$], denoted $v_{lower}$. This move is always feasible for a robot with payload $\rho_{\text{init}}$. Moreover, the closer $\rho_{lower}$ is to $\rho_{\text{init}}$, the higher the quality of the resulting path.
    
    \item CPD($\rho_{upper}$): Similarly, we extract the first move $v_{upper}$ from CPD $(\rho_{upper}$)[$v_i, p_i$]. However, since $\rho_{upper} \geq \rho_{\text{init}}$, $v_{upper}$ may violate slope constraints applicable to $\rho_{\text{init}}$. If a violation occurs, we discard $v_{upper}$ and consider the extraction unsuccessful.
    
\end{itemize}

Our key intuition is that if both CPDs yield the same first move (i.e., $v_{lower} = v_{upper}$), this move is likely to belong to the optimal path $ep(\rho_{\text{init}},v_i, p_i)$. However, this is not always guaranteed due to possible detours or path deviations under different payload conditions. To accommodate this, we take the union of $v_{lower}$ and $v_{upper}$ as the successor set from $v_i$ to $p_i$. Consequently, this process yields at most two successors; if $v_{lower} = v_{upper}$, only one successor is generated.
We apply the same approach to generate successors from $v_j$ to $t$. Thus, each side of the search node $(\{v_i \rightarrow p_i, v_j \rightarrow t\}, f)$ can contribute up to two successors, resulting in at most $2 \times 2 = 4$ combined successors per node. If either side of the search reaches its respective goal, i.e., object $p_i$ or target $t$, the algorithm only considers successors generated from the remaining active side. For instance, for a search node such as $(\{p_i \rightarrow p_i, v_j \rightarrow t\}, f)$, successors are generated only from $v_j$ to $t$. Although this PCPD-based successor generation does not guarantee that the Concurrent CPD Search algorithm remains optimal, our experiments demonstrate that limiting the branching factor to four still achieves near-optimal performance in practice.
 
 \begin{figure}[t]
 \centering
 \includegraphics[width=\columnwidth]{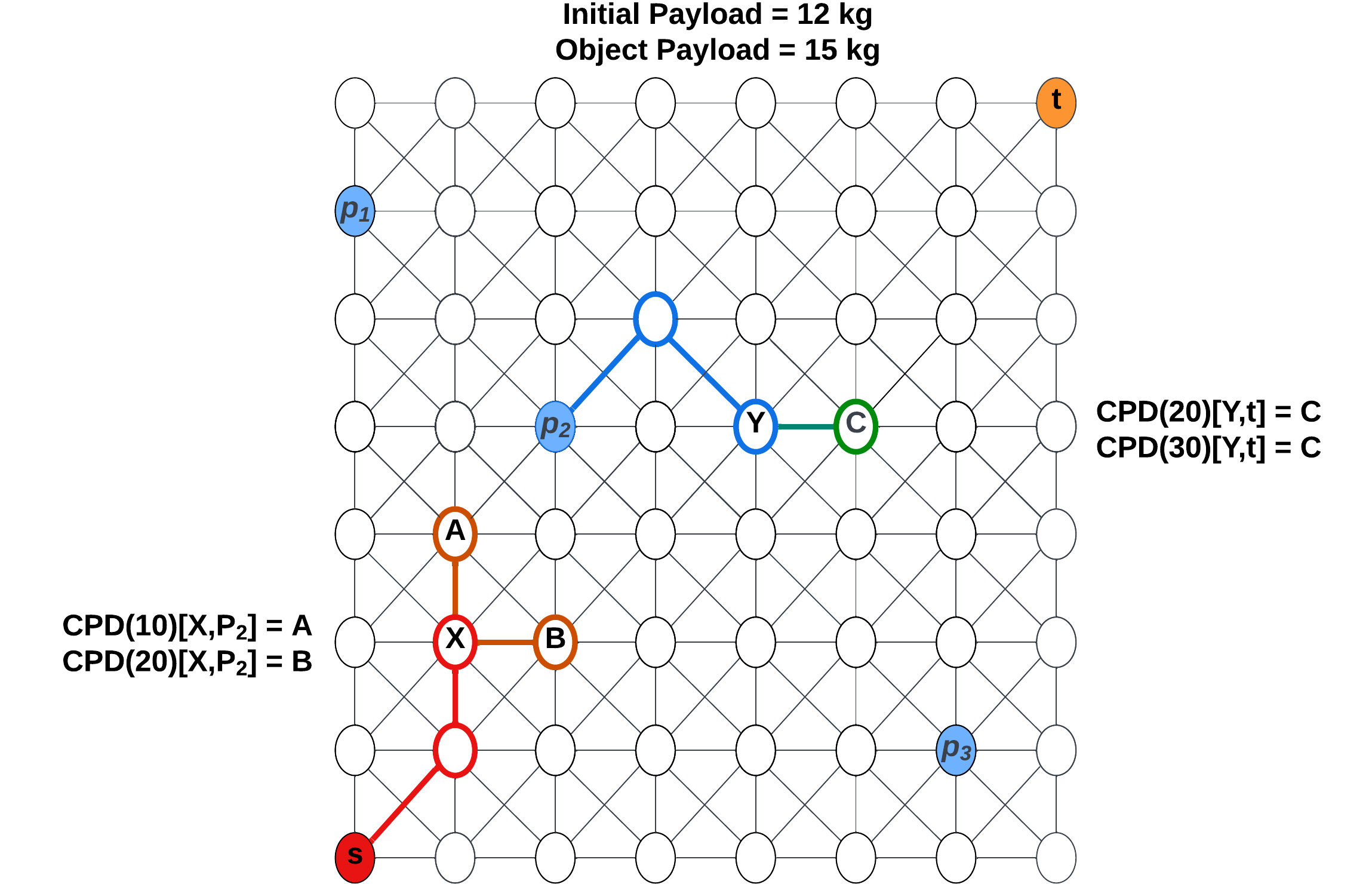}
 \caption{Illustration of successor generation when CPD($\rho_{\text{lower}}$) and CPD($\rho_{\text{upper}}$) suggest two different first moves from the current node $X$ on the $s$–$p_2$ path, while suggesting the same first move from node $Y$ on the $p_2$–$t$ path. This results in two possible combination of successors being pushed to the queue $CQ[p_2]$.
}
   \captionsetup{belowskip=0pt} 
 \label{fig:example}
 \end{figure}
\begin{example}
Consider the toy example illustrated in Fig.~\ref{fig:example}, which includes three pickup points: $\{p_1, p_2, p_3\}$, with an initial payload $\rho_{\text{init}} = 12$ and object payload $\rho_{\text{obj}} = 24$. Our proposed \textit{concurrent PCPD search} adopts a two-level strategy that simultaneously manages multiple Z* searches associated with different pickup points. At the high level, a global priority queue $GQ$ is maintained to coordinate the search across pickup points.
In this example, pickup points $p_1$ and $p_3$ are never expanded due to their high heuristic costs. Specifically, $f(p_1) = h(\rho_{\text{init}}, s, p_1) + h(\rho_{\text{init}}, p_1, t)$ and $f(p_3) = h(\rho_{\text{init}} + \rho_{\text{obj}}, s, p_3) + h(\rho_{\text{init}} + \rho_{\text{obj}}, p_3, t)$, both of which are significantly larger than $f(p_2)$, making $p_2$ the only viable candidate for expansion. The figure demonstrates the successor generation process in the child queue $CQ[p_2]$. The search proceeds from the source node $s$ to the pickup point $p_2$, and simultaneously from $p_2$ to the target $t$, with expanded search nodes shown in red and blue, respectively. Next, we describe how the low-level search generates successors for the search node $(\{X \rightarrow p_2, Y \rightarrow t\}, f)$ using our proposed PCPD-based successor generation technique.

 On the path from $s$ to $p_2$, given $\rho_{\text{init}} = 12$kg, the algorithm queries the PCPD with payload ranges $\rho_{\text{lower}} = 10$ and $\rho_{\text{upper}} = 20$, as defined in Table~\ref{tab:build_time}. At the current search node $X$, it queries both $\text{CPD}(\rho_{\text{lower}})$ and $\text{CPD}(\rho_{\text{upper}})$, where $\text{CPD}(\rho_{\text{lower}})[X, p_2] = A$ and $\text{CPD}(\rho_{\text{upper}})[X, p_2] = B$. Since the CPDs suggest different next moves, two successors are generated: $A$ and $B$.

On the path from $p_2$ to $t$, the robot’s total payload becomes $\rho_{\text{init}} + \rho_{\text{obj}} = 12 + 15 = 27$kg. Accordingly, the algorithm uses CPDs with $\rho_{\text{lower}} = 20$ and $\rho_{\text{upper}} = 30$. At the current search node $Y$, both $\text{CPD}(\rho_{\text{lower}})[Y, t]$ and $\text{CPD}(\rho_{\text{upper}})[Y, t]$ return the same successor $C$, resulting in only one candidate successor: $C$.

Consequently, the concurrent PCPD search combines successors from both directions, generating two joint successors: $(\{A \rightarrow p_2, C \rightarrow t\}, f)$ and $(\{B \rightarrow p_2, C \rightarrow t\}, f)$. Without leveraging PCPD-based successor generation, the search from each side would generate eight successors (one for each outgoing edge), resulting in $8 \times 8 = 64$ combinations. This combinatorial explosion would significantly slow down the search. Our method avoids this inefficiency by pruning the successor space using PCPD-based guidance.

\end{example}

\section{Experiments and Results}
\label{sec:experiments}
In this section, we present our experimental study. We begin by demonstrating the process of extracting data from real-world terrain representations. We then analyze the build time and memory usage of our proposed data structure, the Payload-Constrained Compressed Path Database (PCPD). Finally, we compare the performance of our concurrent PCPD search against a baseline algorithm, evaluating runtime efficiency and solution quality under varying payloads and different numbers of pickup locations.

\subsection{Experimental Setup}

All experiments were conducted on a machine equipped with a 3.0\,GHz Intel Core i7 processor and 32\,GB of RAM. All algorithms were implemented in C++ and compiled using the \texttt{-O3} optimization flag to ensure high-performance execution.

\textbf{Uneven Terrain}:
To simulate realistic navigation scenarios, we selected a $250\,\mathrm{m}^2$ square area from Fairfax County, Virginia, USA. The corresponding Digital Elevation Model (DEM), shown in Fig.~\ref{fig:mash_graph}, contains approximately 250{,}000 nodes. Each node is connected to its eight immediate neighbors, with edge lengths of at least 1\,meter. The resulting graph contains a total of 1{,}993{,}008 links.

\textbf{AMR Configuration}:
The Autonomous Mobile Robot (AMR) used in our simulation is modeled after the Husky A300 unmanned ground vehicle (UGV)\footnote{\url{https://static.generation-robots.com/media/clearpath-robotics-husky-a300-datasheet.pdf}}, a robust, four-wheel-drive platform designed for outdoor applications. For the simulation, we adopted the following physical parameters: robot mass $m = 80\,\mathrm{kg}$, constant velocity $v = 1\,\mathrm{m/s}$, and maximum power availability $P_{\text{max}} = 819.2\,\mathrm{W}$. The dynamic and static friction coefficients were set to $\mu = 0.5$ and $\mu_s = 1.0$, respectively. Although the Husky A300 is rated for a maximum payload of 100\,$\mathrm{kg}$, preliminary experiments indicated that exceeding 60\,$\mathrm{kg}$ significantly reduced the robot’s ability to ascend steep slopes under the specified frictional conditions. Consequently, we capped the payload at 70\,$\mathrm{kg}$ to maintain feasible and reliable climbing behavior in uneven terrain.

\textbf{Query Settings}:
To evaluate algorithmic performance, we generated 1{,}000 random start--target pairs across the terrain. Additionally, 50 pickup locations were randomly distributed throughout the environment. 
The initial payload $\rho_{\text{init}}$ at the start node $s$ and the object payload $\rho_{obj}$ at each pickup location $p$ were randomly assigned to simulate a wide variety of task configurations.
Each query was executed 10 times; the best and worst runtimes were discarded, and the average of the remaining eight runs was reported.

\begingroup
\setlength{\tabcolsep}{1pt}
\renewcommand{\arraystretch}{1.1}
\begin{table*}[t]
    \centering 
     \caption{Build time in Mins, and memory in MB for each CPD built in PCPDs. }
     \small
\begin{tabular}{c||c|c|||c||c|c}
\toprule

\begin{tabular}[c]{@{}c@{}}\textbf{Payload}\\\textbf{Constraint ($\rho'$)}\end{tabular} &
\begin{tabular}[c]{@{}c@{}}\textbf{Build}\\\textbf{Time} \textbf{(Mins)}\end{tabular} & 
\begin{tabular}[c]{@{}c@{}}\textbf{Memory} \\\textbf{(GB)}\end{tabular} &
\begin{tabular}[c]{@{}c@{}}\textbf{Payload}\\\textbf{Constraint} ($\rho'$)\end{tabular} &
\begin{tabular}[c]{@{}c@{}}\textbf{Build}\\\textbf{Time} \textbf{(Mins)}\end{tabular} & 
\begin{tabular}[c]{@{}c@{}}\textbf{Memory} \\\textbf{(GB)}\end{tabular} \\
\midrule
 0 & 101  & 1.3 & 40  & 99.3 & 1.2\\ \hline
10 & 99.9 & 1.3 & 50  & 102 & 1.1 \\ \hline
20 & 102 & 1.3 & 60 &  107& 1.0\\ \hline
30 & 99.5 & 1.3 & 70  & 103 & 0.81\\ \hline

\bottomrule
\end{tabular}
    \label{tab:build_time}
\end{table*}
\endgroup

\begingroup
\setlength{\tabcolsep}{2pt}
\renewcommand{\arraystretch}{1.1}
\begin{table*}[t]
    \centering 
    \caption{Average runtime and suboptimality of the Concurrent PCPD Search compared to the baseline algorithm, evaluated across various initial and object payload configurations. All runtimes are reported in millisecond (ms). Suboptimality is measured as the relative difference from the optimal solution.}

\begin{tabular}{c || c || c || c || c}
\toprule
\multirow{2}{*}{\shortstack{\textbf{Initial} \\ \textbf{Payload} \\ ($\rho_{\text{init}}$)}} &
\multirow{2}{*}{\shortstack{\textbf{Object} \\ \textbf{Payload} \\ ($\rho_{\text{obj}}$)}} &
\textbf{Baseline Algorithm} &
\multicolumn{2}{c}{\shortstack{\textbf{Concurrent} \\ \textbf{PCPD Search}}} \\
\cline{3-5}
& &
\shortstack{\textbf{Avg. Runtime} \\ \textbf{(in ms)}} &
\shortstack{\textbf{Avg. Runtime} \\ \textbf{(in ms)}} &
\shortstack{\textbf{Avg. Suboptimality}} \\

\midrule
4   & 20  & 1716.620 & 6.87362  & 0.00024     \\ \hline
25  & 30  & 1542.683 & 6.97867  & 0.00558     \\ \hline
8   & 46  & 2331.900 & 5.23601  & 0.00512     \\ \hline
6   & 26  & 1775.604 & 6.08912  & 0.00035     \\ \hline
29  & 30  & 1759.997  & 6.29816  & 0.00933      \\ \hline
32  & 24  & 1822.389  & 5.40086  & 0.00694     \\ \hline
45  & 8   & 7099.998  & 7.16672  & 0.01065      \\ \hline
11  & 26  & 1448.384  & 5.76546  & 0.00063     \\ \hline
22  & 19  & 1614.446  & 5.70319  & 0.00072     \\ \hline
9   & 20  & 1803.242  & 5.59241  & 0.00043    \\ \hline

\bottomrule
\end{tabular}
    \label{tab:runtime_comparison}
\end{table*}
\endgroup

\begin{figure*}[t]
\begin{subfigure}{1\textwidth}
  \centering
  \includegraphics[trim=0 140 0 140, clip, width=0.5\linewidth]{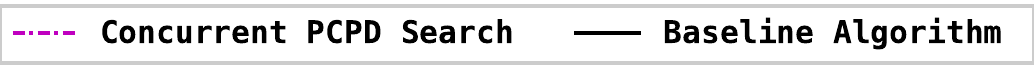}

\end{subfigure}

\begin{subfigure}{.329\textwidth}
  \centering
  
    \includegraphics[width=\linewidth, height=4.2cm]{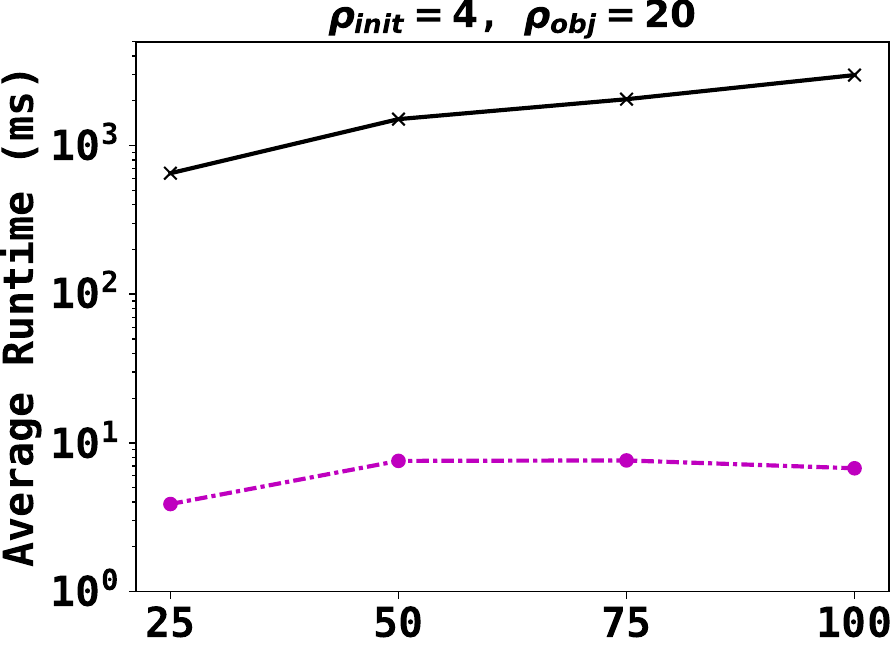}  
  
  \label{fig:dp_4_24}
\end{subfigure}
\begin{subfigure}{.329\textwidth}
  \centering
  
  \includegraphics[width=\linewidth, height=4.2cm]{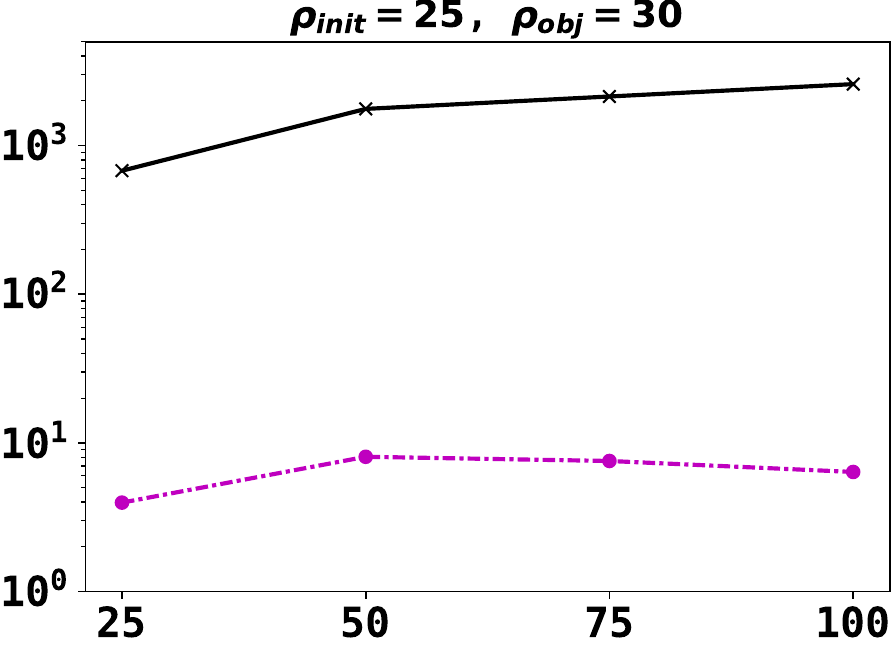}  
  
  \label{fig:dp_25_55}
\end{subfigure}
\begin{subfigure}{.329\textwidth}
  \centering

  \includegraphics[width=\linewidth, height=4.2cm]{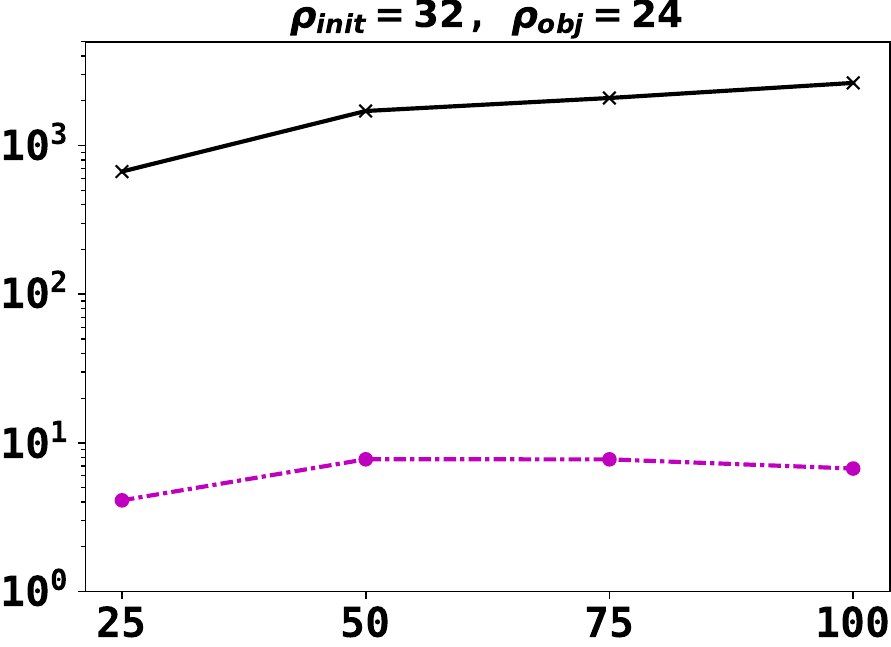}  
  
  \label{fig:dp_32_56}
\end{subfigure}

\caption{
Runtime comparison between Concurrent PCPD Search and the Baseline Algorithm for three different initial and object payloads setting.
Each subplot corresponds to a different configuration of initial payloads ($\rho_{\text{init}}$) and object payloads ($\rho_{\text{obj}}$). The x-axis represents the number of pickup points, and the y-axis  shows the average runtime in milliseconds (ms).
}
\label{fig:runtime_comparison}
\end{figure*}

\subsection{Experiment Results}

\subsubsection{Preprocessing cost.}
Table~\ref{tab:build_time} presents the time (in minutes) and memory usage (in GB) required to construct each CPD$(\rho')$ in the PCPD. As mentioned earlier, we limited the robot's payload to 70 $\mathrm{kg}$ to ensure feasible climbing behavior. Accordingly, we divided the payload range into eight buckets, resulting in eight CPD$(\rho')$ instances that evenly cover the range from 0 to 70 $\mathrm{kg}$, with increments of 10 $\mathrm{kg}$ (i.e., $\rho' = 0, 10, 20, \ldots, 70$).
From the table, we observe that as the payload constraint $\rho'$ increases, the memory usage of each CPD$(\rho')$ decreases. This occurs because a higher payload constraint restricts the robot’s ability to traverse steep slopes, as described in Equation~(\ref{eq:maxAngle}). Consequently, this limitation reduces the number of valid outgoing edges (i.e., feasible first moves) for each vertex. Fewer options lead to greater compression during preprocessing, thereby reducing the overall memory cost.
In addition to that, we observe that the preprocessing time for each CPD$(\rho')$ is typically higher than that reported for other CPD variants~\cite{fast_cpd}.
This is because each CPD$(\rho')$ is constructed by performing a complete modified Dijkstra search from every vertex to generate the first-move table for minimum-energy paths under the payload constraint $\rho'$. This modified Dijkstra is more computationally expensive than the standard version because, at each node expansion, it must validate each outgoing edge $E_{v_iv_j}$ by checking its slope $\theta(v_i,v_j)$ to ensure traversability under the given payload. Additionally, it has to compute the energy cost for each valid edge. 
Overall, constructing each CPD$(\rho')$ takes approximately 100 minutes and consumes around 1 GB of memory. The total memory requirement for all eight CPDs ranges from 8 to 10 GB, which fits well within the capacity of modern systems. Although the preprocessing is moderately intensive, PCPD are generated offline and remain in memory throughout query processing.

\subsubsection{Query Performance.}
Table~\ref{tab:runtime_comparison} compares the runtime performance of our proposed concurrent PCPD search with the baseline algorithm across various randomly generated payload settings. We report the average runtime in milliseconds for both approaches. Since our proposed concurrent PCPD search does not guarantee finding the optimal solution, we measure the suboptimality of its solutions relative to the optimal solutions found by the baseline algorithm. Assuming the path found by concurrent PCPD search is $P$, the suboptimality is calculated as:
\[
\text{subopt}(\rho, s, t) = \frac{ |P|_E - |ep(\rho, s, t)|_E}{ |ep(\rho, s, t)|_E}
\]

We conducted evaluations across 10 different payload configurations, each defined by a unique pair of initial and pickup payload values. As shown in Table~\ref{tab:runtime_comparison}, 
our proposed concurrent PCPD search consistently outperforms the baseline approach by more than two orders of magnitude. This substantial performance gain is primarily attributed to our two-level search framework. The high-level search effectively prunes pickup points $p_i$ that are energetically costly to reach from both the source $s$ and target $t$, based on heuristic estimations. The low-level search then explores the remaining promising pickup points by concurrently performing searches from $s$ to $p_i$ and from $p_i$ to $t$. A key factor contributing to this efficiency is the use of PCPD, which significantly reduces the branching factor of the low-level search, limiting the number of successors to at most four, thereby keeping the search computationally efficient.

In addition to demonstrating superior query runtime performance, Table~\ref{tab:runtime_comparison} reports the average suboptimality of the proposed concurrent PCPD search. The results indicate that our method consistently produces near-optimal solutions, with average suboptimality below 1\% across all payload settings. This validates our PCPD-based successor generation strategy, which restricts the branching factor to a maximum of four successors per node. By querying both $CPD(\rho_{lower})$ and $CPD(\rho_{upper})$, we generate first-move candidates that closely approximate the optimal path. When both CPDs suggest the same move, it is highly likely to be optimal; otherwise, we include both candidates to maintain robustness.
Although this pruning strategy does not guarantee optimality, it significantly reduces the branching factor while preserving solution quality. Our experimental results confirm that this trade-off is effective in practice, enabling efficient search with minimal loss of optimality.

\subsubsection{Effect on Number of Pickup Points}
To further evaluate the runtime performance of our algorithm, we investigate how it scales with an increasing number of pickup points. Figure~\ref{fig:runtime_comparison} presents the average runtime of the proposed concurrent PCPD search compared to a baseline algorithm, as the number of pickup points increases from 25 to 50, 75, and 100. We consider three different payload configurations: (1) $\rho_{\text{init}} = 4$, $\rho_{\text{obj}} = 20$; (2) $\rho_{\text{init}} = 25$, $\rho_{\text{obj}} = 30$; and (3) $\rho_{\text{init}} = 32$, $\rho_{\text{obj}} = 24$. Note that the y-axis is plotted on a logarithmic scale.

The results clearly show that the average runtime of the baseline algorithm increases significantly with the number of pickup points. This is expected, as it must iterate over each candidate pickup point $p_i$, performing two independent Z* searches, one from $s$ to $p_i$ and another from $p_i$ to $t$, without any form of early termination. In contrast, the runtime of our proposed concurrent PCPD search remains stable, even as the number of pickup points increases. This efficiency is primarily attributed to the high-level search strategy, which prioritizes promising pickup points based on heuristic estimates. As a result, unpromising candidates, such as those far away or at locations associated with high energy consumption (e.g., hilltops), are effectively pruned without evaluation. Further performance gains are achieved through our PCPD-based successor generation strategy. By querying both $CPD(\rho_{\text{lower}})$ and $CPD(\rho_{\text{upper}})$, the algorithm identifies accurate first-move candidates and restricts the branching factor to at most four successors per node, significantly reducing per-node expansion costs.
Overall, the concurrent PCPD search consistently outperforms the baseline algorithm by more than two orders of magnitude across all three payload settings, demonstrating superior scalability and efficiency.

\section{Conclusions and Future Work}
\label{sec:conclusion}

In this paper, we introduce the Object-Pickup Minimum Energy Path Problem (OMEPP), a novel formulation addressing the practical challenge of energy-efficient path planning for Autonomous Mobile Robots (AMRs) tasked with picking up an object while traveling between a start point $s$ and a target $t$ across uneven terrain. To solve OMEPP, we first present a baseline algorithm based on the Z* search, which guarantees optimality but incurs high computational costs due to independent repeated searches for each potential pickup location.
To overcome this limitation, we propose a concurrent search strategy leveraging a Payload-Constrained Compressed Path Database (PCPD), an extension of the traditional Compressed Path Database (CPD) that precomputes and stores the first move of energy-efficient paths under different payload constraints. Our concurrent PCPD search algorithm explores multiple pickup options simultaneously and significantly reduces the number of successor nodes generated during the search, resulting in substantial computational savings while maintaining near-optimal solution quality.
Extensive experiments on real-world terrain data demonstrate that our concurrent PCPD search achieves up to two orders of magnitude speedup over the baseline algorithm, with suboptimality consistently below 1\%. Moreover, our results show that the concurrent PCPD search approach scales effectively with increasing numbers of pickup points, making it a viable solution for large-scale deployments.

A promising direction for future research is to extend the OMEPP formulation to scenarios involving multiple objects. In such cases, the robot may be required to transport several objects either in a single trip—subject to cumulative payload constraints—or through multiple trips. This extension introduces additional complexity in planning, particularly in optimizing object pickup sequences and balancing energy consumption across varying payload configurations. Developing efficient algorithms to handle such multi-object variants could significantly enhance the applicability of our approach to more general logistics and autonomous delivery tasks.

%
%
%
%

\bibliographystyle{splncs04}
\bibliography{references}

\begin{thebibliography}{10}
\providecommand{\url}[1]{\texttt{#1}}
\providecommand{\urlprefix}{URL }
\providecommand{\doi}[1]{https://doi.org/#1}

\bibitem{aksamentov2020approach}
Aksamentov, E., Zakharov, K., Tolopilo, D., Usina, E.: Approach to robotic mobile platform path planning upon analysis of aerial imaging data. In: Proceedings of 15th International Conference on Electromechanics and Robotics" Zavalishin's Readings" ER (ZR) 2020, Ufa, Russia, 15--18 April 2020. pp. 93--103. Springer (2020)

\bibitem{alverhed2024autonomous}
Alverhed, E., Hellgren, S., Isaksson, H., Olsson, L., Palmqvist, H., Flod{\'e}n, J.: Autonomous last-mile delivery robots: a literature review. European Transport Research Review  \textbf{16}(1), ~4 (2024)

\bibitem{BabakanoFSC24}
Babakano, F., Fahmin, A., Shen, B., Cheema, M.A.: Time-efficient path planning algorithm for mobile robots on uneven terrain. In: Databases Theory and Applications - 35th Australasian Database Conference, {ADC} 2024, Gold Coast, QLD, Australia, December 16-18, 2024, Proceedings. vol. 15449, pp. 279--292. Springer (2024)

\bibitem{babakano2025time}
Babakano, F., Fahmin, A., Shen, B., Cheema, M.A.: Time-efficient path planning algorithm for mobile robots on uneven terrain. In: Australasian Database Conference. pp. 279--292. Springer (2025)

\bibitem{botea11}
Botea, A.: Ultra-fast optimal pathfinding without runtime search. In: Proceedings of the Seventh {AAAI} Conference on Artificial Intelligence and Interactive Digital Entertainment, {AIIDE} 2011, October 10-14, 2011, Stanford, California, {USA}, 122-127. The {AAAI} Press (2011)

\bibitem{choi2012global}
Choi, S., Park, J., Lim, E., Yu, W.: Global path planning on uneven elevation maps. In: 2012 9th International Conference on Ubiquitous Robots and Ambient Intelligence (URAI). pp. 49--54. IEEE (2012)

\bibitem{ganganath2014finding}
Ganganath, N., Cheng, C.T., Chi, K.T.: Finding energy-efficient paths on uneven terrains. In: 2014 10th France-Japan/8th Europe-Asia Congress on Mecatronics (MECATRONICS2014-Tokyo). pp. 383--388. IEEE (2014)

\bibitem{ganganath2015constraint}
Ganganath, N., Cheng, C.T., Chi, K.T.: A constraint-aware heuristic path planner for finding energy-efficient paths on uneven terrains. IEEE transactions on industrial informatics  \textbf{11}(3),  601--611 (2015)

\bibitem{ganganath2016multiobjective}
Ganganath, N., Cheng, C.T., Chi, K.T.: Multiobjective path planning on uneven terrains based on namoa. In: 2016 IEEE International Symposium on Circuits and Systems (ISCAS). pp. 1846--1849. IEEE (2016)

\bibitem{jeong2024dynamic}
Jeong, J., Moon, I.: Dynamic pickup and delivery problem for autonomous delivery robots in an airport terminal. Computers \& Industrial Engineering  \textbf{196},  110476 (2024)

\bibitem{kyaw2022energy}
Kyaw, P.T., Le, A.V., Veerajagadheswar, P., Elara, M.R., Thu, T.T., Nhan, N.H.K., Van~Duc, P., Vu, M.B.: Energy-efficient path planning of reconfigurable robots in complex environments. IEEE Transactions on Robotics  \textbf{38}(4),  2481--2494 (2022)

\bibitem{liu2013minimizing}
Liu, S., Sun, D.: Minimizing energy consumption of wheeled mobile robots via optimal motion planning. IEEE/ASME Transactions on Mechatronics  \textbf{19}(2),  401--411 (2013)

\bibitem{nguyen2021mobile}
Nguyen, L., Kodagoda, S., Ranasinghe, R., Dissanayake, G.: Mobile robotic sensors for environmental monitoring using gaussian markov random field. Robotica  \textbf{39}(5),  862--884 (2021)

\bibitem{rowe1990optimal}
Rowe, N.C., Ross, R.S.: Optimal grid-free path planning across arbitrarily-contoured terrain with anisotropic friction and gravity effects  (1990)

\bibitem{saad2019energy}
Saad, M., Salameh, A.I., Abdallah, S.: Energy-efficient shortest path planning on uneven terrains: A composite routing metric approach. In: 2019 IEEE International Symposium on Signal Processing and Information Technology (ISSPIT). pp.~1--6. IEEE (2019)

\bibitem{EPS}
Shen, B., Cheema, M.A., Harabor, D., Stuckey, P.J.: Euclidean pathfinding with compressed path databases. In: Proceedings of the Twenty-Ninth International Joint Conference on Artificial Intelligence, {IJCAI} 2020. pp. 4229--4235. ijcai.org (2020)

\bibitem{CHCPD}
Shen, B., Cheema, M.A., Harabor, D.D., Stuckey, P.J.: Contracting and compressing shortest path databases. In: Proceedings of the Thirty-First International Conference on Automated Planning and Scheduling, {ICAPS} 2021, Guangzhou, China (virtual), August 2-13, 2021. pp. 322--330. {AAAI} Press (2021)

\bibitem{fast_cpd}
Strasser, B., Harabor, D., Botea, A.: Fast first-move queries through run-length encoding. In: Proceedings of the Seventh Annual Symposium on Combinatorial Search, {SOCS} 2014, Prague, Czech Republic, 15-17 August 2014. {AAAI} Press (2014)

\bibitem{s-gppc-14}
Sturtevant, N.R., Traish, J.M., Tulip, J.R., Uras, T., Koenig, S., Strasser, B., Botea, A., Harabor, D., Rabin, S.: The {G}rid-based {P}ath {P}lanning {C}ompetition: 2014 {E}ntries and {R}esults. In: Proceedings of the Symposium on Combinatorial Search ({SoCS}). pp. 241--251 (2015)

\bibitem{1391019}
Sun, Z., Reif, J.: On finding energy-minimizing paths on terrains. IEEE Transactions on Robotics  \textbf{21}(1),  102--114 (2005). \doi{10.1109/TRO.2004.837232}

\bibitem{vacariu2004multiagent}
Vacariu, L., Csaba, B.P., Letia, I.A., Fodor, G., Cret, O.: A multiagent cooperative mobile robotics approach for search and rescue missions. IFAC Proceedings Volumes  \textbf{37}(8),  962--967 (2004)

\bibitem{wei2019air}
Wei, M., Isler, V.: Air to ground collaboration for energy-efficient path planning for ground robots. In: 2019 IEEE/RSJ International Conference on Intelligent Robots and Systems (IROS). pp. 1949--1954. IEEE (2019)

\bibitem{wei2020energy}
Wei, M., Isler, V.: Energy-efficient path planning for ground robots by and combining air and ground measurements. In: Conference on Robot Learning. pp. 766--775. PMLR (2020)

\bibitem{wei2024efficient}
Wei, V.J., Wong, R.C.W., Long, C., Mount, D.M., Samet, H.: On efficient shortest path computation on terrain surface: A direction-oriented approach. IEEE Transactions on Knowledge and Data Engineering  (2024)

\bibitem{zakharov2020energy}
Zakharov, K., Saveliev, A., Sivchenko, O.: Energy-efficient path planning algorithm on three-dimensional large-scale terrain maps for mobile robots. In: Interactive Collaborative Robotics: 5th International Conference, ICR 2020, St Petersburg, Russia, October 7-9, 2020, Proceedings 5. pp. 319--330. Springer (2020)

\end{thebibliography}
\end{document}